\documentclass[letterpaper, 10 pt, conference, onecolumn]{ieeeconf}

\IEEEoverridecommandlockouts

\usepackage{graphicx}
\usepackage[dvipsnames]{xcolor}
\usepackage{enumerate}
\usepackage{amssymb}
\usepackage[hidelinks]{hyperref}

\graphicspath{{figs/}}

\def \packagename/{\textsc{VirTooS}}

\title{\LARGE \bf
\packagename/: A ROS~2 -- Unity Virtualization Toolkit \\
for Fleet Management of Autonomous Mobile Robots
}

\author{Andrea~Drudi, Lorenzo~Pichierri, Andrea~Testa, Giuseppe~Notarstefano%
\thanks{This study was carried out within the Space It Up project funded by the Italian Space Agency, ASI, and the Ministry of University and Research, MUR, under contract n. 2024-5-E.0 - CUP n. I53D24000060005.}%
\thanks{Department of Electrical, Electronic and Information Engineering, University of Bologna, Bologna, Italy.
\texttt{\{andrea.drudi4, lorenzo.pichierri, a.testa, giuseppe.notarstefano\}@unibo.it}.}%
}

\begin{document}

\maketitle
\thispagestyle{empty}
\pagestyle{empty}

\begin{abstract}

In this paper, we present \packagename/, a Python/C\# toolkit designed to implement fleet-management tasks on teams of Autonomous Mobile Robots (AMRs). \packagename/ leverages the Robot Operating System (ROS) 2 and Unity game engine to provide realistic, scalable virtual experiments in a mixed-reality environment. 
The toolbox allows users to easily generate and customize virtual scenarios for realistic simulations.
Virtual and real sensors as, e.g., LiDARs, can be exploited to map and safely navigate in the mixed-reality environment.
To enable distributed robotics experiments, we propose a set of tailored routines leveraging the \textsc{ChoiRbot} framework.
As a motivating example, we show a set of experiments for task assignment problems in a virtual environment, allowing seamless interaction among real and virtual robots. 
Moreover, the package comes with a containerized suite to easily deploy it on different machines.
The source code will be made publicly available on GitHub.
\\
\\
Video: \url{https://youtu.be/l6Fs17AwUYE}

\end{abstract}


\section{Introduction}

In recent years, multi-robot systems have attracted significant interest due to their ability to cooperatively address complex tasks in a distributed and scalable manner.
This emerging technology has been increasingly adopted in several domains where the presence of multiple robots is necessary.
Swarm robotics techniques~\cite{hamann2018swarm}, which leverage decentralized control strategies, have proven effective in addressing coordination challenges in large-scale multi-robot systems.
Representative applications include industrial settings~\cite{logothetis2021}, e.g., to handle operations in large-scale indoor environments~\cite{unhelkar2018mobile}, possibly shared with humans~\cite{molina2024iliad}.
The space domain is another relevant application area for multi-robot systems, where robots must operate under severe environmental conditions and communication constraints.
Recent works have explored distributed coordination strategies for satellite swarm control~\cite{liu2018survey} and for cooperative lunar exploration~\cite{de2024multi}.
To address these application frameworks, several works have proposed models, cooperative (possibly distributed) algorithms, and software tools to control and optimize multi-robot systems, see~\cite{keppler2024multi,schwager2024distributedpart1,schwager2024distributedpart2,testa2025tutorial}.

\paragraph*{Related Work} 
We organize the literature relevant to our work in three main parts: (i) ROS~2 frameworks and tools for multi-robot systems, (ii) toolboxes in specific domains, (iii) game engine and mixed-reality tools for robotics.

The Robot Operating System (ROS)~2~\cite{macenski2022robot, macenski2023desks} has become a widely adopted framework for developing applications with autonomous mobile robots.
Existing toolboxes have been proposed for controlling fleets of cooperative mobile robots.
For instance, \textsc{ChoiRbot} \cite{testa2021choirbot} has been proposed as a full-stack ROS~2 toolkit for distributed multi-robot systems, while \textsc{ROS2swarm} \cite{kaiser2022ros2swarm} introduced a ROS~2 framework providing ready-to-use behavioral primitives for robot swarms.
Similarly, \textsc{CrazyChoir} \cite{pichierri2023crazychoir} developed a ROS~2 toolbox for Crazyflie nano-quadrotors.
In \cite{sanchez2024swarm}, the performance of ROS~2-based swarm systems has been experimentally evaluated, highlighting communication and scalability challenges.
In \cite{heuer2024benchmarking}, a ROS~2 benchmark framework, leveraging Gazebo, is proposed to evaluate multi-robot coordination algorithms in realistic, unstructured human-shared environments.

Beyond these general-purpose frameworks, recent studies have investigated ROS~2 in domain-specific multi-robot applications.
In industrial settings, \cite{yumbla2025open} proposes a ROS~2-based multi-robot framework for collaborative environments, whereas \cite{park2020real} provides an experimental assessment of the real-time communication performance of ROS~2 in multi-agent robotic systems.
The works presented in~\cite{erHos2019ros2,erHos2020development} instead focus on the coordination of collaborative robots and other equipment in industrial automation systems, rather than specifically addressing fleets of autonomous mobile robots.
Beyond terrestrial applications, ROS~2 is also being extended to the space domain through Space ROS~\cite{probe2023spaceros}, while Koch et al.~\cite{koch2024slam} present a ROS~2-based SLAM system for the SCOUT planetary cave rover.
Together, these works demonstrate the versatility of ROS~2 across heterogeneous robotic domains and motivate the development of tools for supervising, evaluating, and validating distributed multi-robot systems.

When real deployments are costly, risky, or difficult to access, high-fidelity simulation provides an effective complement to physical experiments.
In this direction, Unity-based environments have been integrated with ROS to enable immersive teleoperation~\cite{whitney2018rosreality}, realistic mobile-robot simulation~\cite{platt2022comparative}, and mixed-reality or digital-twin validation frameworks~\cite{maffettone2024mixed,kwon2025real}.
To the best of our knowledge, however, none of these approaches provides a ROS~2~--~Unity toolkit explicitly tailored to mixed-reality fleet management of AMRs executing distributed coordination algorithms.

\paragraph*{Contributions}
In this paper, we introduce \packagename/, a ROS~2~--~Unity toolkit designed to implement fleet-management tasks on teams of autonomous mobile robots within virtual and mixed-reality environments.
The main core of \packagename/ is written leveraging the ROS~2 framework and Python 3, plus a set of scripts in C\# for the integration with Unity. 
Moreover, the modular structure of the toolbox allows users to modify and extend it according to their needs. 
\packagename/ is built upon the \textsc{ChoiRbot}  toolbox as the primary fleet management system and Unity as the main virtualization engine.
Each robot is characterized by a set of independent modules specifying: $i)$ high-level management functionalities for decision-making, $ii)$ trajectory generation, planning, and control tools, $iii)$ extended functionalities to exchange data with the Unity environment.
In this way, the toolbox allows users to implement photorealistic, virtualized environments in which the robots can navigate.
Sensor measurements reflect the Unity environment.
This enables the execution of typical operations required in complex indoor environments, such as mapping.
To this end, \packagename/ integrates state-of-the-art functionalities provided by the \texttt{ros2-slam-toolbox}~\cite{macenski2021slam}.
To enable safe navigation without collision with other robots or obstacles, each robot performs local replanning while reaching the designated locations.
This is done leveraging the~\texttt{Nav2}~\cite{macenski2020marathon} package.
Finally, to support distributed computing, \packagename/ employs a containerized architecture that ensures scalability and adaptability across diverse robotic platforms.
Its capabilities are demonstrated through virtual and mixed-reality experiments on task-assignment problems, thus showcasing its potential in bridging the gap between algorithmic research and deployment in realistic multi-robot scenarios.

\paragraph*{Organization}
The paper unfolds as follows.
In Section~\ref{sec:architecture}, we detail the architecture of \packagename/, where we present its three main software layers: (i) the optimization-based management system, (ii) the single-robot control, navigation, and planning system, and (iii) the Unity-based interface for virtual sensing and mixed-reality simulations.
Section~\ref{sec:unity-interface} details the Unity~--~Gazebo Virtualization Interface library, which enables seamless integration between real and virtual robots. 
Use-case implementation, virtual and real experiments are provided in Section~\ref{sec:experiments}.

\section{Architecture Overview}
\label{sec:architecture}

\begin{figure}[!htbp]
    \centering
    \includegraphics[width=0.45\textwidth]{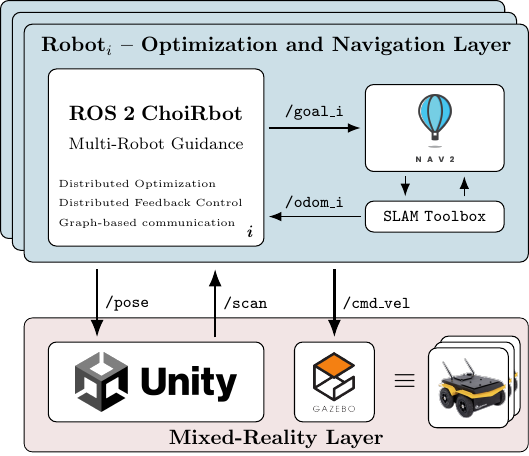}
    \caption{Software architecture of \packagename/ for cooperative robotics in complex indoor environments.}
    \label{unity:fig:architecture}
\end{figure}

The toolbox has a modular structure that divides the software into three main layers: (i) the (optimization-based) team management system, (ii) the single-robot control, navigation, and planning system, and (iii) the Unity-based interface for virtual sensing and mixed-reality simulation.
By coupling these layers, users can implement complex cooperative robotic scenarios by designing their own optimization algorithms, control laws, and virtual environments.
In particular, the Unity-based interface allows users to test and validate their cooperative strategies in realistic virtual environments. Here, robots can navigate and interact within complex environments, simulating real sensor measurements.
This virtualization allows the robots to perform typical operations required in autonomous mobile robot systems, such as mapping and localization.
The toolbox is designed to enable mixed-reality interaction among physical and virtual robots. Unity is leveraged to rapidly generate complex scenarios for testing large robotic fleets in environments that are impractical to reproduce in reality.
Collision-free navigation among virtual and real robots and obstacles is ensured through a layered planning and control architecture.
To meet the demands of distributed computing, this toolbox includes containerized mechanisms offered by Docker, which allows the user to run the software on different operating systems, enabling rapid prototyping on various robotic platforms.
A sketch of the layered software architecture is provided in Fig.~\ref{unity:fig:architecture}.

\subsection{Optimization-based Management System}
In multi-robot systems, fleet behavior is governed by a decision-making layer that typically includes autonomous supervisory systems assigning tasks based on the overall system state and mission objectives.  
Considering the growing complexity of the tasks (e.g., for space missions) and the increasing size of the fleets (e.g., in industrial applications), centralized decision-making architectures become difficult to scale and are being replaced or complemented by distributed approaches.
For these reasons, we built the optimization-based management system on top of \textsc{ChoiRbot}, a toolbox that gives the possibility to implement distributed optimization algorithms to solve complex decision-making scenarios as, e.g., task allocation or vehicle routing problems.
In this layer, each robot is considered as an independent cyber-physical agent that is able to communicate and exchange information with its neighbors and perform local computations to solve its own instances of the optimization problem.
Once the decision-making process is completed, the resulting commands are passed to a lower-level robot controller.

\subsection{Single-Robot Control, Navigation and Planning System}
The second layer of the toolbox is the single-robot control and navigation system.  
This layer is in charge of taking the high-level decisions provided by the optimization-based management system and computing suitable control inputs for the robots.
To facilitate the distribution of the toolbox, we provide plug-and-play solutions that leverage already available and well-established ROS~2 packages for navigation and control, such as \texttt{Nav2}.  
We provide a set of functionalities that adapt the single-robot routines provided by \texttt{Nav2} to the multi-robot scenario, allowing the robots to perform local (and global) replanning and navigation in a shared environment.
With the same objective, we provide multi-robot functionalities to adopt the \texttt{SLAM Toolbox}, allowing the robots to perform SLAM in the shared environment, without the overlapping of ROS~2 routines.

\subsection{Unity~--~Gazebo Virtualization Interface}
\begin{figure}[!htbp]
    \centering
    \includegraphics[width=0.5\textwidth]{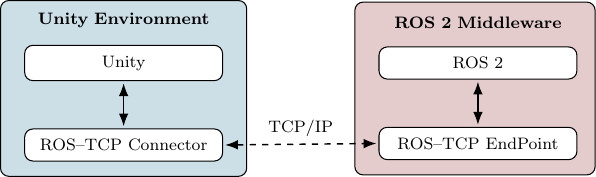}
    \caption{Graphical representation of the communication scheme between Unity and ROS~2.}
    \label{unity:fig:unity-ros}
\end{figure}

To provide a complex simulation environment that supports virtual interactions with (real) robots, we use Unity as the primary virtualization engine in our toolbox.
By providing a set of custom scripts, we replicate an indoor environment where robots can navigate and interact with virtual objects in real time.  
Unity generates the virtual environment, including robots and objects, while Gazebo is employed as the physics engine to simulate the robot dynamics.
This combination is chosen to ensure seamless integration between real and virtual robots, enabling users to leverage existing ROS~2 packages and plugins that rely on Gazebo as their main physics engine.
Communication between Unity and ROS~2 (detailed in Section~\ref{sec:unity-interface}) is facilitated by the \texttt{ROS-TCP-Connector} Unity package, which establishes a TCP/IP connection between Unity and ROS, enabling real-time communication between the two systems.
Acting as the Unity-side component, \texttt{ROS-TCP-Connector} initiates and manages communication by creating a client that can publish, subscribe, and interact with ROS topics, services, and actions.
Its counterpart, \texttt{ROS-TCP-Endpoint}, operates as a ROS node that handles the server-side networking, allowing Unity to send and receive data within the ROS ecosystem (see Fig.~\ref{unity:fig:unity-ros}).
This setup allows the exchange of sensor data and control commands between the robots and both the virtual and real environments.

\section{Unity-based Interface for Virtual Sensing and Mixed-Reality Simulations}
\label{sec:unity-interface}

In this section, we motivate the choice of Unity~\cite{murray2020building} as the main virtualization engine for the toolbox.
Unity is a versatile 3D development platform to create realistic simulations and visualizations, making it ideal for virtualizing environments for multi-robot systems.
Known for its powerful graphics and physics engines, Unity enables the development of dynamic, interactive environments where robots can navigate and interact with objects in real time.

In Unity, core components are \texttt{GameObjects}, \texttt{Prefabs}, and \texttt{Scripts}.
\texttt{GameObjects} are those entities that represent all the virtual objects in a scene, such as robots, humans, environmental features, etc.
\texttt{Prefabs} are reusable templates for \texttt{GameObjects} that, in these applications, enable replication and configuration of robotic models and other assets, such as obstacles, shelves, and functional systems.
For instance, in this toolbox, we use \texttt{Prefabs} to collect all the components, geometric features, settings, and plugins related to a robot (as, e.g., the Jackal ground mobile robot), allowing easy replication in multi-robot scenarios.
\texttt{Scripts} provide custom behaviors to \texttt{GameObjects} through code, allowing developers to control robot dynamics, animations, and interactions with the simulated environment.
In our case, some of the scripts are used to control the communication interface with ROS~2.

Together, these tools enable Unity to be employed as an interactive, high-fidelity simulation front-end, while ROS manages backend robotic processes such as sensor data handling, path planning, and control algorithms.
This setup allows Unity to simulate complex environments and robotic behaviors under ROS control, enhancing the realism and utility of virtual testing and validation in multi-robot systems.
To describe the Unity-based interface, we focus on two main capabilities: (i) the dynamic generation of parametric environments, and (ii) the integration of robot models and virtual sensors for ROS~2-based experiments.

\subsubsection{Dynamic Generation of Virtual Environments}
\begin{figure}[!htbp]
    \centering
    \includegraphics[width=.35\textwidth]{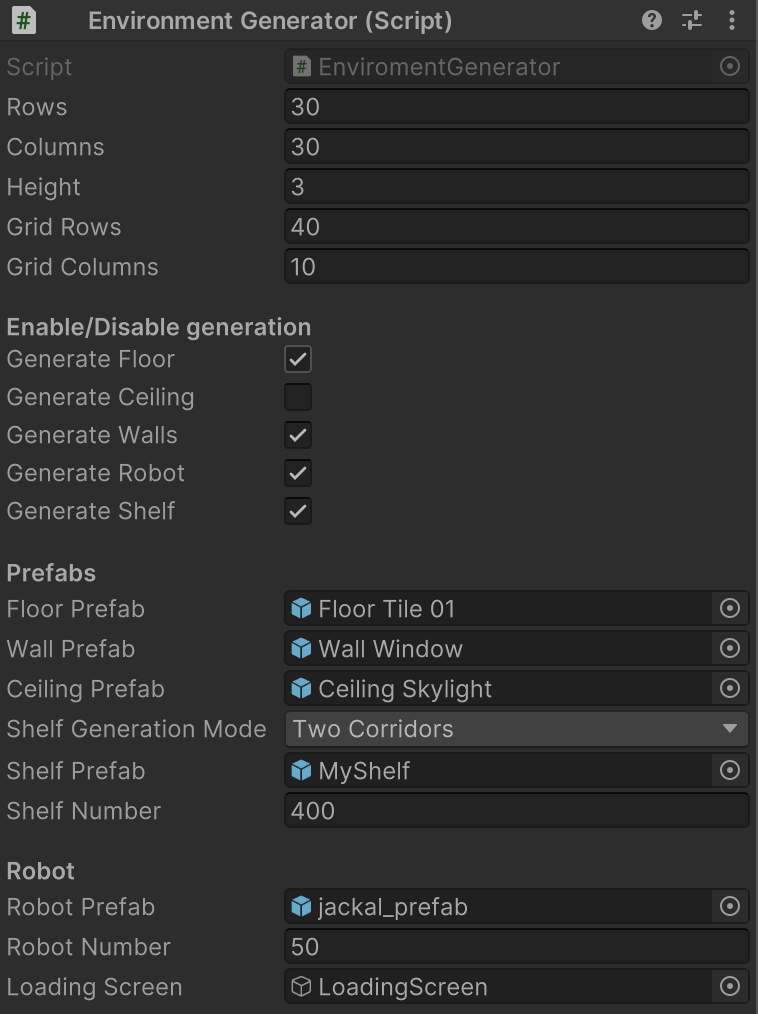}
    \caption{Snapshot of the Unity interface for the dynamic generation of virtual environments. The last menu (i.e., \texttt{Robot}) allows the user to personalize the robot settings to be used in the simulation.}
    \label{unity:fig:environment_generator}
\end{figure}

\begin{figure}[!htbp]
    \centering
    \includegraphics[width=.45\textwidth]{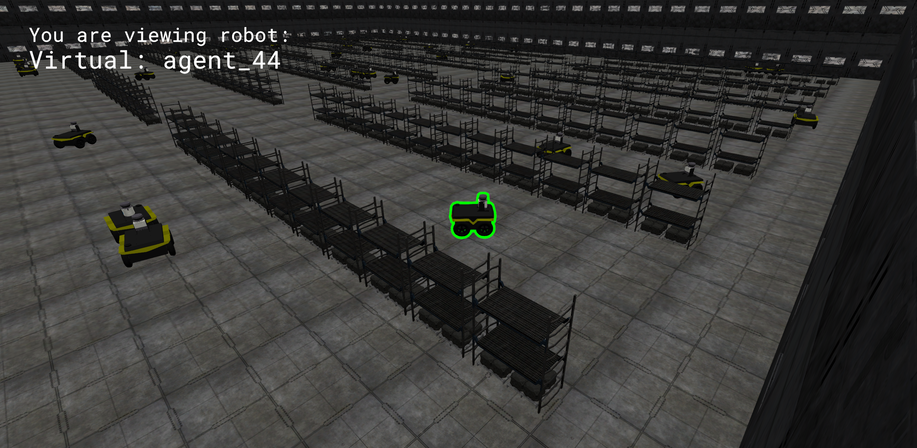}
    \caption{Example of an environment dynamically generated in Unity. In this setup, 50 Jackal ground robots are navigating in a $30\,\mathrm{m}\times30\,\mathrm{m}$ virtual indoor environment.}
    \label{unity:fig:environment}
\end{figure}

A flexible and parametric routine has been developed to build virtual indoor environments, enabling users to customize aspects such as floor layout, walls, shelves, and lighting conditions.
In this dynamic generation, we set the floor tiles in a matrix pattern that can be adjusted through configuration parameters, while walls and ceilings adapt to the dimensions of the floor, creating an enclosed environment.
Objects (e.g., shelves) are placed on a grid corresponding to the ROS~2 reference frame, allowing robots to interact with them on well-defined paths, thus enhancing the realism and functionality of virtual experiments.
The Unity Interface is designed to allow users to easily configure the environment, as shown in Fig.~\ref{unity:fig:environment_generator}.
Here, the user can set the size of the room, the size and the positioning mode of the objects (as, e.g., \texttt{All Rows}, \texttt{Two Corridors}, or \texttt{Random}), and the number of robots.
Also, the user can select the \texttt{Prefab} to be used for the robot and all the other elements in the environment (as, e.g., tiles, walls, and shelves).
The result of an indoor environment setting in Unity is depicted in Fig.~\ref{unity:fig:environment}.

\subsubsection{Robot Integration in Unity}
Robots are integrated in Unity using real-world models that replicate visual and functional aspects. 
In this work, we propose the Jackal robot model, imported from a URDF (Unified Robot Description Format) file through the URDF Importer\footnote{\url{https://github.com/Unity-Technologies/URDF-Importer}} provided by Unity. 

\begin{figure}[!htbp]
    \centering
    \includegraphics[width=.4\textwidth]{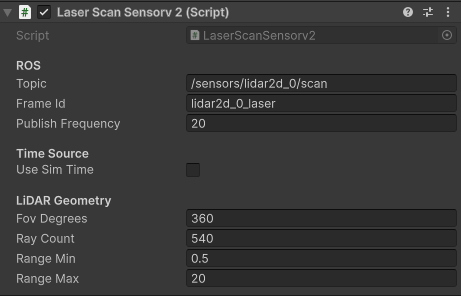}
    \caption{Snapshot of the Unity interface for the configuration of the laser scanner parameters. }
    \label{unity:fig:laser_menu}
\end{figure}

An important aspect of integration is the transformation of coordinate systems. 
Unity follows the RUF (Right-Up-Forward) coordinate convention, whereas Gazebo, employed as a physics engine, uses the FLU (Forward-Left-Up) convention.  
We provide a set of custom scripts that: (i) adjust the orientation of the robot, ensuring consistency between Unity and ROS~2, (ii) manage the exchange of messages, providing Unity with the updated poses and ROS~2 with consistent data coming from the virtualized sensors.  
The laser scanner sensors can be configured through the Unity interface as shown in Fig.~\ref{unity:fig:laser_menu}.
This integration bridges the gap between real and virtual robots, allowing the user to deploy mixed-reality experiments in a controlled, realistic environment.

\section{Mixed-Reality Experiments}
\label{sec:experiments}
In this section, we provide a set of experiments to showcase the functionalities of \packagename/.  
We consider three different scenarios in which a set of virtual and/or real mobile robots have to perform tasks in a virtualized indoor facility.  
The first scenario considers a fleet of Jackal robots solving a Dynamic Task Assignment (DTA)  problem~\cite{burger2012distributed}.  
The second demonstrates the mixed-reality functionalities of \packagename/ by considering a Jackal robot navigating in our laboratory arena while sensing a virtual environment built in Unity.  
Finally, the third experiment combines the first two scenarios and addresses a DTA problem in a mixed-reality environment, where one of the Jackal robots navigates in the physical world while the remaining robots operate in the virtual environment.

\paragraph*{Experimental Setup}
In our experiment, the Jackal robot communicates with the workstation leveraging the well-known ROS~2 publisher-subscriber protocol, which relies on a standard 5 GHz Wi-Fi network.  
The real and virtual experiments have been dockerized and run on Ubuntu 24.04. 
The workstation employed is equipped with an AMD Ryzen 7 7800X3D processor and an Nvidia GeForce RTX 5090 GPU.  
The ROS~2 version used in our experiment is Jazzy Jalisco, while Unity is \texttt{6000.0.74f1}.  
During the mixed-reality experiments, the pose of the robots (i.e., position and orientation) is provided by the Vicon MoCap System.  
A second workstation (Intel i9-9900K, Windows 11) handles the Vicon Tracker 3.10 software.

\subsection{Scenario 1 -- Dynamic Task Assignment in Virtual Indoor Environments with Obstacles}\label{subsec:scenario1}
In this scenario, we assume a dynamically evolving set of tasks that robots must perform cooperatively in a distributed fashion, leveraging their communication capabilities.  
This setup can be modeled as a DTA problem, a well-known combinatorial optimization problem in multi-robot systems.  
The solution of this distributed optimization problem is handled by the optimization-based management system.  
Once each robot determines its task assignment, the single-robot control and navigation system manages the task planning and execution.  
A task is considered accomplished when the robot reaches a specific location in the facility.  
As in real-world applications, we address a dynamic assignment problem where new tasks are introduced as others are completed.  
Consequently, the robots must continuously re-optimize and adjust their local plans in response to new data.

In this virtual experiment, we simulate a team of $N=4$ Clearpath Jackal mobile robots navigating a virtual indoor environment provided by the third main layer of the toolbox: the \textit{Unity--Gazebo Virtualization Interface}.  
Specifically, we use the \texttt{Nav2} package to control the Jackal robots and perform local replanning based on laser scanner measurements, while \texttt{ros2-slam-toolbox} is used to build and update the map of the shared environment.
The interconnection between the simulated robots in Gazebo and their mirrored counterparts in Unity is managed by the \texttt{ROS-TCP-Connector} package, allowing the robots to interact seamlessly with the virtual environment.  
Laser scanner measurements are generated by custom scripts in the Unity suite, while pose information is forwarded by Gazebo.

\begin{figure}[!htbp]
    \centering
    \includegraphics[width=.5\textwidth]{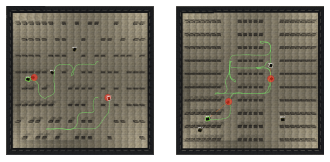}
        \caption{Examples of Task Assignment experiments with robots navigating in two different settings: on the left, randomly placed shelves, while on the right, shelves placed in racks.}
    \label{unity:fig:example_ta}
\end{figure}

\begin{figure}[!htbp]
    \centering
    \includegraphics[width=.5\textwidth]{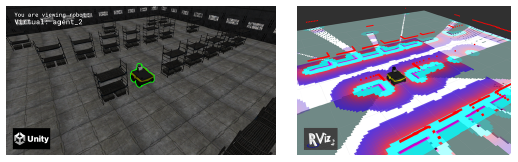}
    \caption{A Jackal mobile robot navigating in a virtual environment generated in Unity (left) while mapping the surroundings (right).}
    \label{unity:fig:robot_pov}
\end{figure}

Fig.~\ref{unity:fig:example_ta} presents snapshots from two experiments where the robots service tasks in different settings.  
Fig.~\ref{unity:fig:robot_pov} illustrates the virtual experiment from the perspective of a robot navigating within the Unity environment, mapping its surroundings, and avoiding collisions with obstacles and other robots.
A video of the experiment is available as supplementary material.

\subsection{Scenario 2 -- Real Robot in Virtual Environment}

In this experiment, we use a Jackal robot navigating in our laboratory while sensing a virtual environment built with the \textit{Unity--Gazebo Virtualization Interface}.  
The real-world robot pose is obtained using a Vicon MoCap system to track its position and orientation, with data acquired through the \texttt{ros2-vicon-receiver} package\footnote{\url{https://github.com/OPT4SMART/ros2-vicon-receiver}}.
This setup ensures that the Jackal pose is continuously updated in real time, providing precise information about its location and orientation.  
The pose data is then transmitted to the \textit{Unity--Gazebo Virtualization Interface} through the \texttt{ROS-TCP-Connector} package, enabling the virtual robot to replicate the physical robot's movements accurately.
On the other hand, laser measurements from the virtual environment are sent to the \textit{Single-Robot Control, Navigation, and Planning System}, allowing the physical robot to perceive and interact with the virtual environment.
This allows for real-time mapping of its surroundings while ensuring collision avoidance with the virtual obstacles.

\begin{figure}[!htbp]
    \centering
    \includegraphics[width=.5\textwidth]{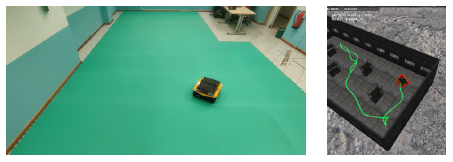}
    \caption{A side-by-side view of a physical Jackal in our laboratory (left) and the virtual twin in the Unity-based simulation (right) navigating in the environment.}
    \label{unity:fig:digital_twin}
\end{figure}

Fig.~\ref{unity:fig:digital_twin} presents a snapshot depicting the Jackal navigating in the real world while sensing the virtual environment.  
To highlight the mixed-reality interconnection demonstrated in this experiment, the robot is tasked with reaching a series of goal poses within the virtual environment in the presence of (virtual) obstacles.

\subsection{Scenario 3 -- Mixed-Reality Experiment}
The objective of this last experiment is to showcase how \packagename/ can be utilized in cases where access to a complete fleet of physical robots is limited. 
By employing this approach, users can validate motion control algorithms and sensor configurations on a physical robot while concurrently testing cooperative algorithms for fleet management in a high-fidelity virtual environment. 
This dual capability provides a robotic platform for evaluating complex multi-robot coordination strategies, even in resource-constrained scenarios, advancing the development of scalable, cooperative robotic systems.

In this scenario, we employ a fleet of $N=4$ Jackal robots, where one robot navigates in the real world, while the other ones operate in the virtual environment.
In the experiment, the robots solve an instance of the Dynamic Task Assignment problem to cooperatively service a set of tasks. 
Moreover, the physical robot is mirrored in the virtual world, thus interacting with the virtual environment and avoiding collisions with the virtual robots.

\begin{figure}[!htbp]
    \centering
    \includegraphics[width=.45\textwidth]{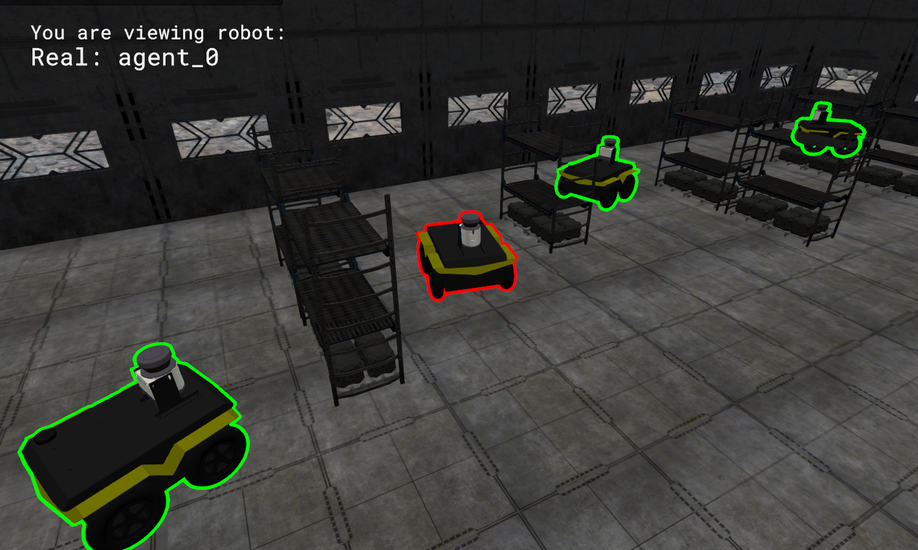}
    \caption{A snapshot from the mixed-reality environment. The red robot is the virtual twin of the real Jackal, while the green robots are the fully virtual robots.}
    \label{unity:fig:world_exp}
\end{figure}

The experiment evolves in the same way as the one described in Section~\ref{subsec:scenario1}, with two main differences:

\begin{enumerate}[\it i)]
    \item \texttt{$Robot_{1,R}$} is a physical robot,
    \item the environment is a $4\,\mathrm{m} \times 8\,\mathrm{m}$ room (see Fig.~\ref{unity:fig:world_exp}), to facilitate the navigation of the physical robot in our laboratory.
\end{enumerate}

\begin{figure}[!htbp]
    \centering
    \includegraphics[width=.9\textwidth]{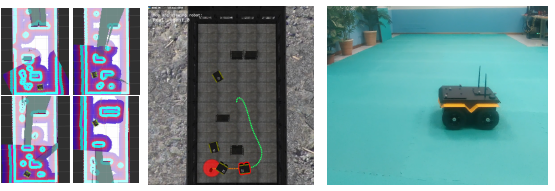}
    \caption{Snapshot of the mixed-reality experiment at $t = 25\,\mathrm{s}$. The red robot represents the virtual twin of the physical Jackal.}
    \label{unity:fig:mixed_real}
\end{figure}

\begin{figure}[!htbp]
    \centering
    \includegraphics[width=.9\textwidth]{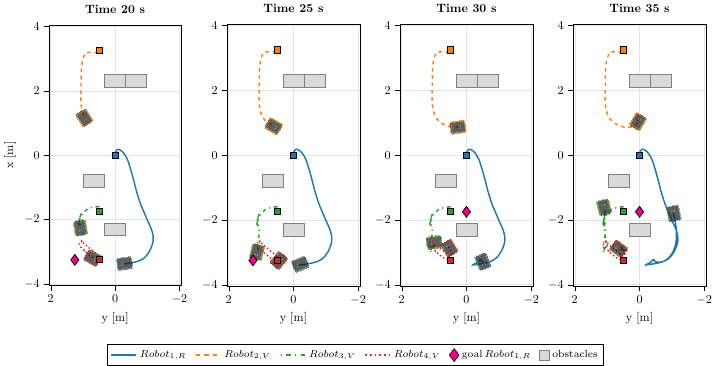}
    \caption{Snapshots of the real ($Robot_{1,R}$) and virtual ($Robot_{i,V}$) robots' trajectories during the mixed-reality experiment at different time instants. The square marker (\scalebox{0.7}{$\square$}) represents the initial robot position.}
    \label{unity:fig:traj_snapshots}
\end{figure}

Fig.~\ref{unity:fig:mixed_real} and \ref{unity:fig:traj_snapshots} show a sequence of snapshots from the mixed-reality experiment, highlighting the evolution of the robot trajectories during the navigation task.
The physical robot, denoted as $Robot_{1,R}$, operates in the real environment while interacting with the virtual robots through the shared mixed-reality framework.
The remaining agents, namely $Robot_{2,V}$, $Robot_{3,V}$, and $Robot_{4,V}$, are simulated within the virtual environment. 
The figures illustrate how the physical robot successfully navigates toward the assigned goal (high-level task allocation) while safely avoiding both the static obstacles and the virtual robots (local navigation stack). 
Between $20\,\mathrm{s}$ and $30\,\mathrm{s}$, the robots exhibit limited motion since the local collision avoidance behavior prevents safe progress toward the assigned goals in the constrained environment. 
Once the fleet management layer updates the target assignments, the robots resume navigation toward the goal.
A video of the experiment is available as supplementary material.

\paragraph*{Other Applications}
\packagename/ is not limited to the previous examples. 
It can be used to implement several multi-robot algorithms in virtual environments.
For instance, it can be used to test and validate cooperative mapping and localization algorithms, or distributed optimization algorithms for multi-robot systems, as, e.g., the Pick-up and Delivery Vehicle Routing Problem~\cite{camisa2022multi}.
Moreover, thanks to the photorealistic environment provided by Unity, it can be used to test and validate vision-based navigation algorithms~\cite{desouza2002vision}, and, for example, to implement novel fleet management techniques allowing safe human-robot interaction in realistic environments, see, e.g.,~\cite{molina2024iliad}.
\packagename/ can also support the prototyping and validation of multi-robot coordination strategies in exploration applications where harsh environments and communication constraints make extensive real-world testing difficult, as, e.g., in space robotics~\cite{liu2018survey,de2024multi}.

\section{Conclusions}
In this paper, we introduced \packagename/, a Python and C\# toolkit designed to facilitate the implementation of fleet-management tasks for teams of mobile robots operating in complex environments. 
By leveraging the capabilities of the Robot Operating System (ROS)~2 and Unity frameworks, \packagename/ enables high-fidelity virtual experiments in mixed-reality scenarios. 
The package provides an accessible interface for users to design custom virtual environments, implement cooperative algorithms, and exploit virtual onboard sensors.
A key feature of the proposed toolkit is its containerized architecture, which supports seamless deployment across diverse robotic platforms. 
Moreover, the integration of state-of-the-art toolboxes, such as, e.g., \texttt{Nav2} and \texttt{SLAM Toolbox}, extends its applicability.

\bibliographystyle{IEEEtran}
\bibliography{IEEEabrv,mybibfile}

@STRING{IEEE_J_CST        = "{IEEE} Trans. Control Syst. Technol."}

@STRING{IEEE_J_PAMI       = "{IEEE} Trans. Pattern Anal. Mach. Intell."}

@STRING{IEEE_J_RAL        = "{IEEE} Robot. Autom. Lett."}

@STRING{IEEE_J_RO         = "{IEEE} Trans. Robot."}

@STRING{IEEE_J_PROC       = "Proc. {IEEE}"}

@STRING{IEEE_O_ACC        = "{IEEE} Access"}

@STRING{IEEE_M_RA         = "{IEEE} Robot. Autom. Mag."}

@article{burger2012distributed,
  title     = "A distributed simplex algorithm for degenerate linear programs and multi-agent assignments",
  author    = "B{\"u}rger, Mathias and Notarstefano, Giuseppe and Bullo, Francesco and Allg{\"o}wer, Frank",
  journal   = "Automatica",
  volume    = "48",
  number    = "9",
  pages     = "2298--2304",
  year      = "2012",
  publisher = "Elsevier"
}

@article{camisa2022multi,
  title     = "Multi-robot pickup and delivery via distributed resource allocation",
  author    = "Camisa, Andrea and Testa, Andrea and Notarstefano, Giuseppe",
  journal   = IEEE_J_RO,
  volume    = "39",
  number    = "2",
  pages     = "1106--1118",
  year      = "2022",
  publisher = "IEEE"
}

@inproceedings{de2024multi,
  title="Multi-agent autonomy for space exploration on the cadre lunar technology demonstration",
  author="de la Croix, Jean-Pierre and Rossi, Federico and Brockers, Roland and Aguilar, Dustin and Albee, Keenan and Boroson, Elizabeth and Cauligi, Abhishek and Delaune, Jeff and Hewitt, Robert and Kogan, Dima and others",
  booktitle="2024 IEEE Aerospace Conference",
  pages="1--14",
  year="2024",
  organization="IEEE"
}

@article{desouza2002vision,
  title="Vision for mobile robot navigation: A survey",
  author="DeSouza, Guilherme N and Kak, Avinash C",
  journal = IEEE_J_PAMI,
  volume="24",
  number="2",
  pages="237--267",
  year="2002",
  publisher="IEEE"
}

@article{erHos2019ros2,
  title     = "A {ROS2} based communication architecture for control in collaborative and intelligent automation systems",
  author    = "Er{\H{o}}s, Endre and Dahl, Martin and Bengtsson, Kristofer and Hanna, Atieh and Falkman, Petter",
  journal   = "Procedia Manufacturing",
  volume    = "38",
  pages     = "349--357",
  year      = "2019",
  publisher = "Elsevier"
}

@incollection{erHos2020development,
  title="Development of an industry 4.0 demonstrator using sequence planner and {ROS2}",
  author="Er{\H{o}}s, Endre and Dahl, Martin and Hanna, Atieh and G{\"o}tvall, Per-Lage and Falkman, Petter and Bengtsson, Kristofer",
  booktitle="Robot Operating System (ROS) The Complete Reference (Volume 5)",
  pages="3--29",
  year="2020",
  publisher="Springer"
}

@book{hamann2018swarm,
  title     = "Swarm robotics: A formal approach",
  author    = "Hamann, Heiko",
  volume    = "221",
  year      = "2018",
  publisher = "Springer"
}

@inproceedings{heuer2024benchmarking,
  title        = "Benchmarking Multi-robot coordination in realistic, unstructured human-shared environments",
  author       = "Heuer, Lukas and Palmieri, Luigi and Mannucci, Anna and Koenig, Sven and Magnusson, Martin",
  booktitle    = "2024 IEEE International Conference on Robotics and Automation (ICRA)",
  pages        = "14541--14547",
  year         = "2024",
  organization = "IEEE"
}

@inproceedings{kaiser2022ros2swarm,
  title     = "{ROS2SWARM} - a {ROS} 2 package for swarm robot behaviors",
  author    = "Kaiser, Tanja Katharina and Begemann, Marian Johannes and Plattenteich, Tavia and Schilling, Lars and Schildbach, Georg and Hamann, Heiko",
  booktitle = "IEEE International Conference on Robotics and Automation",
  pages     = "6875--6881",
  year      = "2022"
}

@phdthesis{keppler2024multi,
  title       = "Multi-robot Trajectory Coordination for Complex Vehicle Combinations",
  author      = "Keppler, Felix",
  year        = "2024",
  school      = "Technische Universit{\"a}t Dresden",
  type        = "Doctoral dissertation",
  institution = "Fraunhofer-Institut f{\"u}r Verkehrs- und Infrastruktursysteme IVI",
  publisher   = "Fraunhofer Verlag",
  series      = "Wissenschaftliche Schriften des Fraunhofer IVI / Scientific publications by Fraunhofer IVI",
  number      = "1",
  pages       = "123",
  isbn        = "978-3-8396-2052-6",
  doi         = "10.24406/publica-3722",
  url         = "https://www.bookshop.fraunhofer.de/buch/multi-robot-trajectory-coordination-for-complex-vehicle-combinations/256187",
  language    = "English"
}

@inproceedings{koch2024slam,
  author = "Koch, Johannes and Sch{\"u}tt, Manuel and Ryll, Markus",
  year = "2024",
  month = "November",
  pages = "515--521",
  booktitle = "i-SAIRAS 2024 Paper Book",
  title = "SLAM for SCOUT: A {ROS2}-Based Multi-Sensor System for Compliant Rovers",
  url = "https://elib.dlr.de/210182/"
}

@article{kwon2025real,
  author="Kwon, Woojin and Yang, Jieun and Song, Seunghwa and Lee, Jun and Kim, Hyungjung",
  journal = IEEE_O_ACC,
  title="Real-Time Digital-Twin-Based Cobot-Worker Collision Risk Prediction Using Unity, {ROS}, and UWB", 
  year="2025",
  volume="13",
  number="",
  pages="85967-85978",
  doi="10.1109/ACCESS.2025.3569332"
}

@article{liu2018survey,
  title="A survey on formation control of small satellites",
  author="Liu, Guo-Ping and Zhang, Shijie",
  journal = IEEE_J_PROC,
  volume="106",
  number="3",
  pages="440--457",
  year="2018",
  publisher="IEEE"
}

@article{logothetis2021,
  author="Logothetis, Michalis and Karras, George and Alevizos, Konstantinos and Verginis, Christos and Roque, Pedro and Roditakis, Konstantinos and Makris, Alexandros and Garcia, Sergio and Schillinger, Philipp and Di Fava, Alessandro and Pelliccione, Patrizio and Argyros, Antonis and Kyriakopoulos, Kostas and Dimarogonas, Dimos V.",
  journal = IEEE_M_RA,
  title="Efficient Cooperation of Heterogeneous Robotic Agents: A Decentralized Framework", 
  year="2021",
  volume="28",
  number="2",
  pages="74-87",
  doi="10.1109/MRA.2021.3064761"
}

@INPROCEEDINGS{macenski2020marathon,
  author="Macenski, Steve and Martín, Francisco and White, Ruffin and Clavero, Jonatan Ginés",
  booktitle="2020 IEEE/RSJ International Conference on Intelligent Robots and Systems (IROS)", 
  title="The Marathon 2: A Navigation System", 
  year="2020",
  volume="",
  number="",
  pages="2718-2725",
  doi="10.1109/IROS45743.2020.9341207",
  url="https://github.com/ros-navigation/navigation2"
}

@article{macenski2021slam,
  title   = "SLAM Toolbox: SLAM for the dynamic world",
  author  = "Macenski, Steve and Jambrecic, Ivona",
  journal = "Journal of Open Source Software",
  volume  = "6",
  number  = "61",
  pages   = "2783",
  year    = "2021"
}

@article{macenski2022robot,
  author = "Steven Macenski  and Tully Foote  and Brian Gerkey  and Chris Lalancette  and William Woodall ",
  title = "Robot Operating System 2: Design, architecture, and uses in the wild",
  journal = "Science Robotics",
  volume = "7",
  number = "66",
  pages = "eabm6074",
  year = "2022",
  doi = "10.1126/scirobotics.abm6074",
  URL = "https://www.science.org/doi/abs/10.1126/scirobotics.abm6074",
  eprint = "https://www.science.org/doi/pdf/10.1126/scirobotics.abm6074"
}

@article{macenski2023desks,
  title="From the desks of {ROS} maintainers: A survey of modern \& capable mobile robotics algorithms in the robot operating system 2",
  author="Macenski, Steve and Moore, Tom and Lu, David V and Merzlyakov, Alexey and Ferguson, Michael",
  journal="Robotics and Autonomous Systems",
  volume="168",
  pages="104493",
  year="2023",
  publisher="Elsevier"
}

@article{maffettone2024mixed,
  title="Mixed reality environment and high-dimensional continuification control for swarm robotics",
  author="Maffettone, Gian Carlo and Liguori, Lorenzo and Palermo, Eduardo and Di Bernardo, Mario and Porfiri, Maurizio",
  journal = IEEE_J_CST,
  volume="32",
  number="6",
  pages="2484--2491",
  year="2024",
  publisher="IEEE"
}

@article{molina2024iliad,
  author="Molina, Sergi and Mannucci, Anna and Magnusson, Martin and Adolfsson, Daniel and Andreasson, Henrik and Hamad, Mazin and Abdolshah, Saeed and Chadalavada, Ravi Teja and Palmieri, Luigi and Linder, Timm and Swaminathan, Chittaranjan Srinivas and Kucner, Tomasz Piotr and Hanheide, Marc and Fernandez-Carmona, Manuel and Cielniak, Grzegorz and Duckett, Tom and Pecora, Federico and Bokesand, Simon and Arras, Kai O. and Haddadin, Sami and Lilienthal, Achim J.",
  journal = IEEE_M_RA,
  title="The ILIAD Safety Stack: Human-Aware Infrastructure-Free Navigation of Industrial Mobile Robots", 
  year="2024",
  volume="31",
  number="3",
  pages="48-59",
  doi="10.1109/MRA.2023.3296983"
}

@book{murray2020building,
  title     = "Building virtual reality with unity and steamvr",
  author    = "Murray, Jeff W",
  year      = "2020",
  publisher = "Crc Press",
  doi= "10.1201/9780429295850"
}

@article{park2020real,
  title="Real-time characteristics of {ROS} 2.0 in multiagent robot systems: An empirical study",
  author="Park, Jaeho and Delgado, Raimarius and Choi, Byoung Wook",
  journal = IEEE_O_ACC,
  volume="8",
  pages="154637--154651",
  year="2020",
  publisher="IEEE"
}

@article{pichierri2023crazychoir,
  author="Pichierri, Lorenzo and Testa, Andrea and Notarstefano, Giuseppe",
  journal = IEEE_J_RAL,
  title="{CrazyChoir}: Flying Swarms of Crazyflie Quadrotors in {ROS} 2", 
  year="2023",
  volume="8",
  number="8",
  pages="4713-4720",
  doi="10.1109/LRA.2023.3286814"
}

@article{platt2022comparative,
  title="Comparative analysis of ros-unity3d and ros-gazebo for mobile ground robot simulation",
  author="Platt, Jonathan and Ricks, Kenneth",
  journal="Journal of Intelligent \& Robotic Systems",
  volume="106",
  number="4",
  pages="80",
  year="2022",
  publisher="Springer"
}

@inproceedings{probe2023spaceros,
  title="Space ros: An open-source framework for space robotics and flight software",
  author="Probe, Austin and Oyake, Amalaye and Chambers, S W and Deans, Matthew and Brat, Guillaume and Cramer, Nick B and Kempa, Brian and Roberts, Brian and Hambuchen, Kimberly",
  booktitle="AIAA SciteCH 2023 Forum",
  pages="2709",
  year="2023"
}

@article{sanchez2024swarm,
  author="Castillo-Sánchez, José-Borja and González-Parada, Eva and Cano-García, José-Manuel",
  journal = IEEE_O_ACC,
  title="Swarm Robot Communications in {ROS} 2: An Experimental Study", 
  year="2024",
  volume="12",
  number="",
  pages="142930-142943",
  doi="10.1109/ACCESS.2024.3470254"
}

@article{schwager2024distributedpart1,
  author="Shorinwa, Ola and Halsted, Trevor and Yu, Javier and Schwager, Mac",
  journal = IEEE_M_RA,
  title="Distributed Optimization Methods for Multi-Robot Systems: Part 1—A Tutorial [Tutorial]", 
  year="2024",
  volume="31",
  number="3",
  pages="121-138",
  doi="10.1109/MRA.2024.3358718"
}

@article{schwager2024distributedpart2,
  author="Shorinwa, Ola and Halsted, Trevor and Yu, Javier and Schwager, Mac",
  journal = IEEE_M_RA,
  title="Distributed Optimization Methods for Multi-Robot Systems: Part 2—A Survey", 
  year="2024",
  volume="31",
  number="3",
  pages="154-169",
  doi="10.1109/MRA.2024.3352852"
}

@article{testa2021choirbot,
  title     = "{ChoiRbot}: A {ROS} 2 toolbox for cooperative robotics",
  author    = "Testa, Andrea and Camisa, Andrea and Notarstefano, Giuseppe",
  journal   = IEEE_J_RAL,
  volume    = "6",
  number    = "2",
  pages     = "2714--2720",
  year      = "2021",
  publisher = "IEEE"
}

@article{testa2025tutorial,
  author="Testa, Andrea and Carnevale, Guido and Notarstefano, Giuseppe",
  journal = IEEE_J_PROC,
  title="A Tutorial on Distributed Optimization for Cooperative Robotics: From Setups and Algorithms to Toolboxes and Research Directions", 
  year="2025",
  volume="113",
  number="1",
  pages="40-65",
  doi="10.1109/JPROC.2025.3557698"
}

@article{unhelkar2018mobile,
  title="Mobile robots for moving-floor assembly lines: Design, evaluation, and deployment",
  author="Unhelkar, Vaibhav V and D{\"o}rr, Stefan and Bubeck, Alexander and Lasota, Przemyslaw A and Perez, Jorge and Siu, Ho Chit and Boerkoel, James C and Tyroller, Quirin and Bix, Johannes and Bartscher, Stefan and others",
  journal = IEEE_M_RA,
  volume="25",
  number="2",
  pages="72--81",
  year="2018",
  publisher="IEEE"
}

@inproceedings{whitney2018rosreality,
  title="Ros reality: A virtual reality framework using consumer-grade hardware for ros-enabled robots",
  author="Whitney, David and Rosen, Eric and Ullman, Daniel and Phillips, Elizabeth and Tellex, Stefanie",
  booktitle="2018 IEEE/RSJ International Conference on Intelligent Robots and Systems (IROS)",
  pages="1--9",
  year="2018",
  organization="IEEE"
}

@article{yumbla2025open,
  title="An Open-Source Multi-Robot Framework System for Collaborative Environments Based on {ROS2}",
  author="Yumbla, Francisco and Fajardo-Pruna, Marcelo and Piguave, Anthonny and Ronquillo, Diego and Ortiz, Ricardo and Choi, Jongseong Brad and Diaz, Gabriel and Pa{\~n}eda, Xabiel G and Moon, Hyungpil",
  journal = IEEE_O_ACC,
  volume="13",
  pages="16288--16302",
  year="2025",
  publisher="IEEE"
}

\end{document}